# Measuring an Artificial Intelligence System's Performance on a Verbal IQ Test For Young Children[*]

Stellan Ohlsson[1], Robert H. Sloan[2], György Turán[3][4], Aaron Urasky[3]

**Affiliations and email:**

[1] Department of Psychology, University of Illinois at Chicago, Chicago, IL 60607, stellan@uic.edu.

[2] Department of Computer Science, University of Illinois at Chicago, Chicago, IL 60607, sloan@uic.edu.

[3] Department of Mathematics, Statistics and Computer Science, University of Illinois at Chicago, Chicago, IL 60607, gyt@uic.edu (Turán), aaron.urasky@gmail.com (Urasky).

[4] MTA-SZTE Research Group on Artificial Intelligence, Szeged, Hungary.

**Abstract**

We administered the Verbal IQ (VIQ) part of the Wechsler Preschool and Primary Scale of Intelligence (WPPSI-III) to the ConceptNet 4 AI system. The test questions (e.g., "Why do we shake hands?") were translated into ConceptNet 4 inputs using a combination of the simple natural language processing tools that come with ConceptNet together with short Python programs that we wrote. The question answering used a version of ConceptNet based on spectral methods.

The ConceptNet system scored a WPPSI-III VIQ that is average for a four-year-old child, but below average for 5 to 7 year-olds. Large variations among subtests indicate potential areas of improvement. In particular, results were strongest for the Vocabulary and Similarities subtests, intermediate for the Information subtest, and lowest for the Comprehension and Word Reasoning subtests. Comprehension is the subtest most strongly associated with common sense.

The large variations among subtests and ordinary common sense strongly suggest that the WPPSI-III VIQ results do *not* show that "ConceptNet has the verbal abilities a four-year-old." Rather, children's IQ tests offer one objective metric for the evaluation and comparison of AI systems. Also, this work continues previous research on Psychometric AI.

## 1. Introduction

Russell and Norvig's hugely popular textbook [33] promotes developing devices that *act rationally* as the key goal for Artificial Intelligence (AI). They emphasize that acting rationally sometimes means outperforming humans (e.g., computers have outperformed humans in arithmetic from the very beginning, and computers have outperformed humans in chess in recent years). Of course, acting rationally also means performing at least as well as humans on any task.

How should we measure performance? *Psychometric AI (PAI)* [3, 4] (see also, e.g., [8, 9, 38]) argues that for many areas, psychometric tests developed for humans provide ready-made tools. *Psychometrics* itself is the branch of psychology that studies the objective measurement of human skills and traits, including human intelligence. A psychometric intelligence test has two main components: First, a fixed list of *test items*, each item requiring some recordable response by the testee. The second main component of an intelligence test is the set of norming tables. That is, prior to use, each test item is presented to a large sample of persons drawn from the population for which the test is intended. The testee's answer to a test item is quantified by comparing it to the statistical distribution of answers from the sample population. The testee's intelligence is quantified by aggregating the normed answers to the individual test items. Because different populations will generate different norming data, there are separate tests for different populations. For example, the intelligence of children is measured by a different test than the intelligence of adults. Because of the high development costs, psychometric intelligence tests are commercial, proprietary products.

---

[*] Preliminary reports on this work appeared as a 2-page AAAI 2013 late-breaking development poster and in a workshop [29, 30].

In this paper describe our results using one such test, a Verbal IQ test for young children, to measure some areas where, as far as we know, currently AI systems do not come anywhere close to adult human performance.

There has long been a feeling that there are some general reasoning and/or verbal tasks that average human children can perform but that AI systems cannot as yet perform. This general idea has appeared in the AI literature from the 1950s until today. In particular, in the late 1950s John McCarthy's paper "Programs with Common Sense'' [25] noted that "certain elementary verbal reasoning processes so simple that they can be carried out by any non-feeble minded human have yet to be simulated by machine programs.'' Thirty years later, Davis described common sense for an AI system as "common knowledge about the world that is possessed by every schoolchild and the methods for making obvious inferences from this knowledge" and illustrated this definition with an example "easily understood by five-year-old children" [11]. Roughly twenty years after that, in 2008 Rodney Brooks gave a list of challenges for artificial intelligence [5, 27] that included performing with the language capabilities of a 4-year-old child, and the social understanding of an 8-year-old child.

Our goal in this work is to demonstrate the use of a metric in the spirit of PAI that can help make more precise the tasks where AI systems are currently struggling to meet the performance of 4- to 8-year-old children. We believe a full-blown verbal IQ test is such a metric. In (at least) US schools today, the Wechsler series of IQ tests are the most commonly used IQ tests when a full-blown IQ test is called for, for example, to check for developmental delays. Most portions of the Wechsler IQ tests (and all the verbal portions) consist of open-response questions (i.e., *not* multiple choice) that are administered by a psychologist in a one-on-one session with the child being tested. Some subtests consist of questions that it would be very easy to write a computer program to answer; for example, giving back the definition of a particular word. However, other portions have questions that at least intuitively highlight current difficulty spots for AI systems. For example, consider the following three questions:[1]

1. Why do we wear sunscreen in the summer?
2. Why is it bad to put a knife in your mouth?
3. What should you do if you see a strange grown-up lying in the street?

As far as we know, there are not yet AI question-answering systems that can reliably give good answers to all three of those questions, and yet an average 8-year-old child could.

Incidentally, we also cannot yet use the Web to answer all three of those questions in an automated fashion. For the first question, a Google or Bing query will return a *pointer* to a website answering the question, but it will not return the answer. For the second question, Bing and Google might perhaps include a pointer to the right answer among their top five results, but then again, they might not.[2] For the third question, search engines are unlikely to be of any help.

---

[1] The actual items in the Wechsler IQ tests are all proprietary; all example questions we give have been made up by the authors of this paper.

However, most average seven or eight-year-olds can answer all three questions. Also, all three of those questions are in the general spirit of questions on the Information and/or Comprehension subtests of the Wechsler IQ tests.

In our work, we used the verbal portions of the *Wechsler Preschool and Primary Scale of Intelligence*, Third Edition (WPPSI-III) test, a multi-dimensional IQ test designed to assess the intelligence of children of ages 4–7.25 years, which is widely used to measure the IQ of young children. (We describe this test in more detail below).

We chose the WPPSI-III because we expected some of its subtests to highlight *limitations* of current AI systems, as opposed to some PAI work with other psychometric tests of verbal abilities that highlights the progress that AI systems have made over the decades. There has been a fair amount of work in the past fifteen years on using AI systems for answering multiple-choice verbal questions in the spirit of such tests as the SAT, ACT, and GRE (e.g., [39]). Most of the work takes on analogy questions (e.g., "Hammer is to nail as saw is to (a) wood, (b) screws, (c) tools, (d) toolkit, (e) woodshop"), but synonyms, antonyms, and classification questions have also been considered (e.g., [38]).

The main point of this article is to demonstrate the use of the WPPSI-III as a PAI instrument, but of course we still had to test on some particular AI system or systems. We, like many others before us, hypothesized that to answer some of the questions a large knowledge base would be required. Indeed, this notion goes back at least to Edward Feigenbaum's *knowledge principle* [6, 15]: High performance on intellectual tasks is based on a large knowledge base.

Therefore, as our initial trial of the WPPSI-III as a PAI instrument, we started out trying to test two systems based on large knowledge bases: ConceptNet/AnalogySpace [16, 17, 37] and Cyc [22], two representative systems that have both been around for some time, and are in the public domain. (ConceptNet is fully in the public domain; Cyc is has some versions in the public domain.) In our preliminary investigations we were more successful with ConceptNet 4 than with Cyc, so we worked with ConceptNet 4 for this initial exploration. It turned out in our preliminary investigations that the particular approach we were interested in is more suitable for ConceptNet 4 than for Cyc and therefore we worked with ConceptNet 4 for the remainder of the project (see the end of Section 1.2 for more details).

Section 1.1, which follows, reviews related work, primarily prior studies in which some AI system was applied to test items from some form of intelligence test. Section 1.2 gives an overview of how we conducted the present study, while Section 1.3 provides a preview of our contribution. In Sections 2 and 3 we provide the technical details about the AI system and the intelligence test. In Section 4 we give our methods in detail, and in Section 5 we report our results. We end in Section 6 with our conclusions and some general discussion and speculation.

**1.1 Prior Work**

Legg and Hutter survey several possible approaches to testing AI systems' performance [21], including the use of psychometric tests. This approach is, in fact, almost as old as McCarthy's challenge. In the 1960s, T. D. Evans developed a program that solved multiple-choice visual analogy items taken from an IQ test [14]. He compared the performance of his program to data from humans and found that, depending on the exact version, system performance ranged from slightly below the median score for students in 9th grade to slightly above the median for students in 12th grade. Another example from the 1960s is the Argus program for verbal analogies [32], which "also smack of IQ tests'' [19]. As far as we know, after these 1960s efforts, there was no follow-up work in AI until the 2000s (although there *was* theoretical work on commonsense reasoning; see, e.g., Davis [11] and Mueller [26]).

---
[2] We tried typing, "Why is it bad to put a knife in your mouth" into each of Bing, DuckDuckGo, and Google in December 2014, and did not get any useful answer in the top five hits from any of those search engines.

In 2003, Bringsjord and Schimanski [4] coined the term "psychometric AI (PAI)" and argued for the usefulness of psychometric tests of intelligence as tools for driving the development of AI systems. Since then, multiple works on PAI have appeared. These efforts tend to focus on test items specific to some particular domain of knowledge, such as mechanical comprehension [20]. However, Bringsjord and Schimanski emphasized the need to develop programs that can perform well on tests of general intelligence, and the work we present here takes that approach.

Since 2000, several researchers have tested AI systems on multiple choice questions or questions requiring the insertion of a number or a letter to complete a pattern. Sources of the questions include books for the general public, websites, and also some released versions of some commercial tests, such as the SAT. We briefly discuss several of these works here.

Veale used the taxonomic structure of Wordnet as the basis for a program answering multiple-choice analogy questions from the SAT [39]. Sanghi and Dowe [34] programmed a system that can respond to three types of nonverbal questions They report IQ scores of the program for 12 different "IQ tests," from various books and websites. Their program scored at or slightly above the human average of 100 an all but three of the 12 tests, though it is very unclear whether the tests used were in fact carefully normed. In subsequent articles, Dowe and Hernández-Orallo have argued against this empirical approach [13]. Instead, they advocate a rational approach to measuring the intelligence of an AI system based on theoretical and formal considerations [18].

Turney [38] built a system based on analogies, and was able to use it to obtain reasonably strong results on seven different questions types, including multiple-choice synonym questions from the TOEFL (test of English as a foreign language) and classifying word pairs as either synonyms or antonyms, as well as analogy questions from the SAT.

Wang et al. [40] identified five types of items used in multiple-choice verbal "IQ tests:" Two types of analogies, classification, synonym and antonym items. (The test items came from several books of "IQ test" style quizzes for the general public, such as *The Times Book of IQ Tests*.) They created an item type recognition algorithm by training a machine learning system on a collection of such items. Their system outperformed three alternative systems, and exhibited an average percentage of correct answers at or above the levels achieved by 200 human participants recruited over the Internet via the Mechanical Turk.

The most important difference between those works and our work is that none of those works attempted to answer questions that are even vaguely like those found in the Information, Word Reasoning, and Comprehension subtests of the WPPSI-III. (See Section 3 for more details on the WPPSI-III.) Also, many of the tests labeled "IQ tests" in those papers did not have the careful norming of full-blown commercial IQ tests. Another difference is that the systems described above used solution routines that were either brute force or based on statistical machine learning.

The Winograd Schema Challenge [23] is another metric that has been proposed to highlight verbal problems that are easy for humans but appear to be difficult for AI systems. The task is to resolve pronoun antecedent ambiguity in sentences where the change of one word causes the pronoun to change its antecedent. For example, what does the pronoun "it" refer to in the sentence, "The trophy doesn't fit in the suitcase because it is too big / small." The Winograd Schema is *not* a psychometric, because the questions are designed so that any normally competent speaker of English can easily answer them correctly. Very recently Weston et al. proposed another group of benchmarks for question answering also involving questions that normal human speakers could answer all of without difficulty [41].

## 1.2 Present Work

Our psychometric test is the *Wechsler Preschool and Primary Scale of Intelligence*, Third Edition (WPPSI-III) test, a multi-dimensional IQ test designed to assess the intelligence of children of ages 4–7.25 years. (Some parts of it can be used with children between 2.5 and 4 years old.) It is widely used to measure the IQ of young children. The full WPPSI-III returns a (WPPSI-III) Verbal IQ (VIQ), a (WPPSI-III) Performance IQ (PIQ), and a (WPPSI-III) Full Scale IQ. For this article, we restricted ourselves to only the WPPSI-III VIQ items.

The WPPSI-III, and IQ testing in general, has some particular strengths and some particular weaknesses for use in PAI. One great advantage of the major commercial IQ tests is that there is a huge norming effort involving a great amount of time, money, and test subjects. (Wikipedia reports that 1700 children were used in the norming of the US version of the WPPSI-III.) There are not many instruments that have this degree of norming for human performance. Of those that do, there are even fewer that are targeted at and normed for young children. The adult Wechsler IQ test (the WAIS) and also the SAT, ACT, and GRE for instance, are obviously very carefully normed, but they target the ability level of 16-year-olds to adults.

In our opinion another major advantage of a full-blown VIQ test for PAI is that it highlights areas that are easy for humans and currently appear to still be difficult for AI systems.

A significant disadvantage of the major full-blown commercial IQ tests is that the questions are proprietary, and access to the questions is very tightly guarded. New editions of these tests are created only once every ten to twenty years, and frequently earlier versions remain in limited use for some time, so the questions are simply never released. Therefore, as is the case for all scientific work reporting on IQ testing, we will not report on any specific questions in the WPPSI-III.

Another possible disadvantage is that the scoring of many sections of full-blown IQ tests is somewhat subjective, although the scoring guide that is part of the WPPSI-III makes most of the scoring fairly unambiguous.

Also, there is a long-standing debate regarding the usefulness of intelligence tests for managing societal affairs [31]. However, using intelligence tests as standardized performance measures to make systematic comparisons among AI systems and between machines and humans does not directly engage that debate.

In order to explore whether using such an IQ test was even possible, we first made up our own test items in the general spirit of, but distinct from, those included in the WPPSI-III. We began with preliminary attempts to answer questions in the style of the WPPSI-III using both ConceptNet and Cyc.

Our goal was to evaluate how well a system would perform when used by ordinary computer scientists, not by experts in question answering or by the system's creators. One of our interests is in large knowledge bases, such as ConceptNet, Cyc, or NELL [7], and such knowledge bases must be usable by computer science researchers (and eventually practitioners) in general, not only by their creators. We were at least somewhat successful in using the ConceptNet 4 plus AnalogySpace system. However, we were unable to use Cyc to answer more than a few of our sample questions. This seems to indicate a limitation of Cyc as used by computer scientists not specifically trained in Cyc, but not necessarily an inherent limitation of Cyc as used by experts on Cyc.

In the rest of the paper we only discuss the results obtained for the ConceptNet/AnalogySpace system developed at MIT. We provide a brief overview of that system in Section 2.

**1.3 Summary of Contributions and Results**

The main contributions of our work are as follows: (1) We show how to use an actual Verbal IQ test as a metric as part of the efforts of psychometric AI. This is significant because the IQ test is *the* psychometric used, when necessary, to give a comprehensive view of young children. (2) We highlight the strengths and weaknesses of at least one commonsense AI knowledge base, indicating areas of focus for further research; (3) We demonstrate how strongly entangled the issues of knowledge representation, commonsense reasoning, natural language processing, and question answering are; (4) We give empirical results on ConceptNet's abilities measured by the verbal portion of the WPPSI-III.

One overall conclusion of our work is that ConceptNet has a WPPSI-III VIQ of an average four-year-old child. As discussed later on in more detail, our implementation of query-answering algorithms was simple. The details of ConceptNet's performance raise many interesting issues and challenges concerning how an AI system might obtain a higher score using more sophisticated algorithms. Some of these are briefly discussed at the end of this article.

**2. The AI System**

ConceptNet is an open-source project run by the MIT Common Sense Computing Initiative. It has several components. The Open Mind Common Sense initiative originally acquired a large common-sense knowledge base from web users [35]. This is ConceptNet itself, consisting of triples of the form (<concept1>, *relation*, <concept2>), where *relation* is drawn from a fixed set of about twenty relations, including *IsA*, *Causes*, and *AtLocation*.

More precisely, each entry in ConceptNet 4 consists of two "concepts" and one of the relations, together with either "left" or "right" to show the direction of the relation (e.g., to indicate that "a fawn IsA deer" as opposed to "a deer IsA fawn"). There is also a numerical strength, and a polarity flag. The latter is set in a small minority of cases (3.4 percent) to indicate negation (e.g., polarity could be used to express the assertion that "Penguins are not capable of flying."). There is also a frequency, which we did not use in this work.

AnalogySpace [36, 37] is a concise version of the ConceptNet knowledge base that its creators say is "designed to facilitate reasoning over" ConceptNet [27]. Leaving out assertions that have relatively little support shrinks the number of concepts from about 275,000 to about 22,000 for the English-language version. Additional shrinkage comes from treating the ConceptNet knowledge base as a large but sparse matrix and applying spectral techniques, specifically a truncated singular value decomposition (SVD) to obtain a smaller, denser matrix. This reduced-dimension matrix, which is called AnalogySpace, is claimed to give better, more meaningful descriptions of the knowledge [37]. To be a little more precise, the original matrix has rows that are concepts, and columns that are features, that is, ordered pairs of relations and concepts. The signs of the matrices entries indicate the polarity. For example "the feature vector for 'steering wheel' could have +1 in the position for 'is part of a car', +1 for 'is round', and −1 for 'is alive' " [37]. A truncated SVD is applied to that matrix to give AnalogySpace. For more details, see Speer et al. [37].

In the work reported here, we used the March 2012 joint release of ConceptNet 4 implemented as the Python module conceptnet and AnalogySpace implemented as the Python module divisi2.[3] In this paper "ConceptNet" refers to this combination of AnalogySpace and ConceptNet 4 unless explicitly stated otherwise.

---

[3] AnalogySpace was originally released for ConceptNet 3 and was updated for ConceptNet 4. As of this writing a version 5 of ConceptNet has been released but no corresponding version of AnalogySpace has yet been released.

## 3. The Intelligence Test

The WPPSI-III IQ test is composed of 14 subtests. A complete WPPSI-III IQ assessment consists of a Performance IQ (derived from drawing, puzzle, and memory tasks) and a Verbal IQ (VIQ). Performance and Verbal IQ can be combined into a full-scale IQ score. Each IQ score has a mean of 100 and a standard deviation of 15. We used the five subtests that can combine to yield VIQ scores. VIQ is determined by three of five verbal subtests: *Information*, *Vocabulary*, *Word Reasoning*, *Comprehension*, and *Similarities*. The first three are "core" and the last two are "supplemental." The examiner may choose any three subtests to compute a VIQ score, as long as at least two are core subtests.

For some subtests each item is scored as either 0 (incorrect) or 1 (correct); for some subtests each item is scored as 0 (incorrect), 1 (partial credit), or 2 (fully correct), and for some subtests some of the simpler questions towards the start are scored 0–1 but later questions are scored 0–1–2. Each subtest has its own rule for stopping, such as, "Stop after five items in a row are scored zero." For each subtest, a raw score is obtained by adding up all the points earned for all the items administered before stopping. That raw score is not considered important, however. Raw scores are converted using the testees's age and the WPPSI-III norming tables into a scaled score, a whole number in the range of 1 to 19, which has a mean of 10 and a standard deviation of 3. Normally the scaled rather than raw scores are the subtest scores that psychologists, parents, and educators care about.

As discussed in Section 1.2, the WPPSI-III test items are proprietary, so we will not describe any specific test items. For developing and testing our approach, we made up test items of the same type as, but distinct from, the WPPSI-III items. All examples of test items we give are taken from our own item pool. All *scores* reported in this article were obtained with the actual WPPSI-III items, used under license. We reported on our planned methodology and some preliminary results based only on our own item pool in [28].

Next, we briefly describe each item type for the five verbal subtests.

In a *Vocabulary* item, the testee is asked to articulate the meaning of a common word, using the question frame, "What is ___?", as in "What is a house?" Success on a vocabulary item requires retrieval of the relevant concept definition, and the lack of retrieval of irrelevant concepts.

In an *Information* item, the testee is asked to state the properties, kind, function, cause, origin, consequence, location, or other aspect of some everyday object, event, or process. For example, the testee might be asked, "Where can you find a penguin?"

In a *Similarities* item, two words have to be related using the sentence frame, "Finish what I say. X and Y are both ___", as in "Finish what I say. Pen and pencil are both ___". Performance on a Similarities item requires the retrieval of the two concept definitions plus the ability to find a meaningful overlap between them.

In a *Word Reasoning* item, the task is to identify a concept based on one to three clues. The testee might be told, "You can see through it," as a first clue; if the correct answer is not forthcoming, the testee might be told that, "It is square and you can open it." The processing required by a Word Reasoning items goes beyond retrieval because the testee has to integrate the clues and choose among alternative hypotheses.

Finally, in a *Comprehension* item, the task is to construct an explanation of general principles or social situations in response to a question, usually a why-question. The testee might be asked, "Why do people shake hands?" Performance on a comprehension item requires the construction of an explanation, and so goes beyond retrieval. In some descriptions of the WPPSI-III, the *Comprehension* subtest is described as a test of common sense.

## 4. Method and Some Examples

### 4.1 Formulating Queries to ConceptNet

The VIQ scores we obtained are necessarily a function of both ConceptNet itself *and* our algorithms for administering the items to the system. The VIQ scores also depend on human scoring of the answers received. The proper scoring of actual WPPSI-III items is described in great detail in the materials given to WPPSI-III administrators. For each question the materials provide both a detailed description of what constitutes a correct answer as well as lengthy lists of typical correct and incorrect answers. Incidentally, it is a requirement that the scoring be done by either a licensed clinical psychologist or a university researcher in psychology (in this case Ohlsson).

We wrote short programs in Python to feed each of the five types of verbal test items into ConceptNet. We used the natural language processing tools that come with ConceptNet, and added some additional but minimal natural language processing of our own.

We used our own test items to develop our method and to choose the amount of truncation of the SVD. For the results reported here, we truncated to the first $k = 500$ most significant singular values, but our results were similar for any value of $k$ in the 200 to 600 range. We describe our methodology for the two straight question-answering subtests, Information and Comprehension, in some detail, and then give shorter descriptions for the rest of the subtests, highlighting the new issues they raise.

In general, we attempted to use the ConceptNet system as intended, making a few straightforward adjustments, such as the routine for "what color" questions we describe below. We wrote under 500 lines of Python code in total for all our routines for answering the five different types of subtest questions. All of the code we wrote is available at https://www.dropbox.com/l/JNmvkihhhT47VfdbT7rMiq.

As far as we know, there are not yet any AI systems that could answer all five types of WPPSI-III VIQ items at all successfully if they were presented purely in natural language. For example, consider what an AI question-answering system would be likely to do with, "Finish what I say. Pen and pencil are both ___". So any approach to using a current AI system to attempt the complete VIQ portion of the WPPSI-III necessarily involves writing some code of one's own, and the performance obtained necessarily reflects the combination of the AI system and the additional code. We chose to go the route of writing a minimal amount of extra code.

**Information and Comprehension:** The AnalogySpace materials suggest answering questions by a two-step process:

1. Use ConceptNet's natural language tools to map the input question to a set of concepts that are in ConceptNet. For example, "Why do we put on sunscreen in summer?" is mapped to a set of four ConceptNet concepts: "why", "put sunscreen", "put" "sunscreen", and "summer". "Why is it bad to put a knife in your mouth?" is also mapped to a set of five ConceptNet concepts, namely "why", "bad", "put" "knife", and "mouth". In ConceptNet's terminology, that set is called a category.

2. Take the product of the entire AnalogySpace matrix and the category (treating the category as a column vector) to get a vector of proposed answers. The proposed answers will be "features," which is to say a relation together with a direction and a concept. The direction tells whether the feature is, for example "star *IsA*" or "*IsA* star". Each feature also has a numerical score. Return the highest scored feature.

We gave ConceptNet a bit of help by writing Python scripts to handle the first word of common forms of questions. For "why" we removed the concept "why" before querying AnalogySpace, and we only used answers whose relation was one of *Causes, Desires, UsedFor, HasPrerequisite, CausesDesire, MotivatedByGoal,* or *HasSubevent*. Thus we threw away any proposed answers involving any of ConceptNet's other twenty or so relations, such as *IsA* or *HasA*. For the knife in the mouth question we got the answer "*UsedFor* cut," which we scored as wrong, but is at least vaguely in ballpark. For the sunscreen question we got the dreadful answer "*UsedFor* box." We did get some very good answers to why-questions. For instance, "Why do we have refrigerators in our kitchens" was answered "*UsedFor* store food", and "Why do we shower?" was answered "*Causes* become clean."

Similarly, for where-questions we restricted the answers to involve only the relations *AtLocation* and *NearLocation*, and for questions beginning "What" we restricted the answers to 13 relations that were a better fit for "what". Our made-up Information question, "Where can you find a penguin" got the correct answer "*AtLocation* zoo," although "Where does your food go when you swallow it" got the answer "*AtLocation* refrigerator."

We also looked for a short list of phrases in questions that matched very well to one of ConceptNet's relations. For example, if the question contained "use" or "used" then we required the answer to have the *UsedFor* relation and if the question contained "made of", "make from", or "made out of" we required the answer to have the relation *MadeOf*. We also remove those words from the query passed to AnalogySpace.

We got the idea for those restrictions when our question "What is made out of wood" was converted by the natural language tools to the four concepts "make out", "make" "out", and "wood" and the answer returned concerned the notion of a couple making out. Instead passing only "wood" to AnalogySpace we get "paper *MadeOf*". (It was just our good luck that "paper MadeOf" was the top scoring *MadeOf* relation for wood, instead of, say, "*MadeOf* tree". A more sophisticated question answering routing would try to parse the question more carefully to determine which side of the *MadeOf* relation we need to answer a particular query.)

As we discuss more in Section 5, we also removed one-word concepts that were part of two-word concepts also in the parse returned from ConceptNet's NLP tools after our other processing; thus we removed both 'put' and 'sunscreen' from the NLP-to-concept mapping of "… put on sunscreen in summer" that originally returned ['put', 'sunscreen', 'put sunscreen', 'summer'] and both 'shake' and 'hands' from the NLP-to-concept mapping of "Why do people shake hands?" that originally included all three of 'shake', 'hand', and 'shake hand'.

We also wrote two special subroutines, one for "What color is" and one for "How many" questions. We initially asked the question "What color is the sky?" and were pleased to get the answer "*IsA* Blue" and did not realize we needed any special processing. However, it later turned out that almost all "What color" questions were being answered "blue," presumably because "blue" has a higher association with "color" in AnalogySpace than, for example, "white." But of course "blue" is a terrible answer to the question, "What color is snow?"

The processing of questions beginning "What color is/are", "What is the color of", and "How many" was the same as for other questions with the following modifications. First, for the color questions we removed the concept "color" from the set of concepts passed to AnalogySpace. Second, we filtered the concepts returned as answers, and returned only concepts $c$ having a sufficiently high score for ($c$, IsA, color) or ($c$, IsA, Number), according to the type of question. Our threshold score was 95 percent of the score of the lowest-scoring color or number from a set of relatively common colors and numbers that we tried. (Indigo turns out to have the lowest IsA-color score in ConceptNet of any of black, white, and the ROYGBIV colors.) We also excluded as answers the concepts 'color', 'number', 'person', 'yourself', 'many', 'part', 'organ', and 'much', each of which turned out to have a surprisingly high score for at least one of IsA-color or IsA-number.

An overview of the entire process, showing how the question "Why do we put on sunscreen in summer?" is processed (and eventually yields the very bad answer "Sunscreen UsedFor Box") is given in Figure 1.

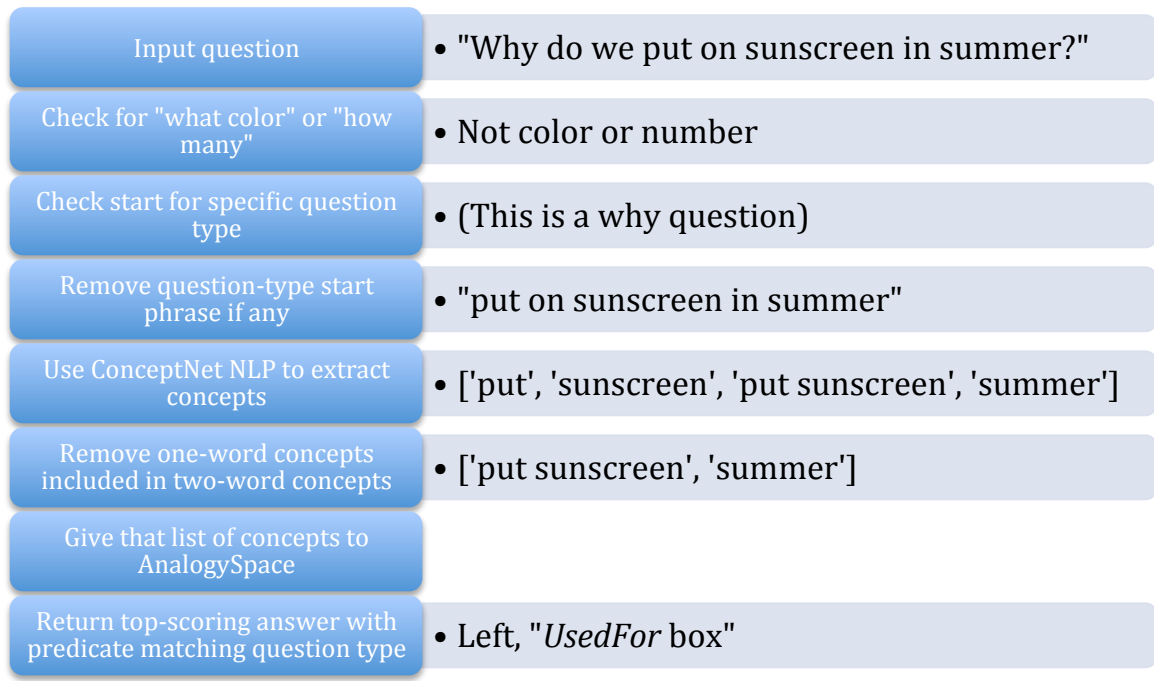

**Fig. 1**. Example of processing flow for Information and Comprehension questions. For why-questions, the predicates we allowed were 'HasPrerequisite','CausesDesire', 'MotivatedByGoal', 'Causes', 'UsedFor', 'HasSubevent', and 'Desires'. Notice that the answer is (very) wrong.

**Word Reasoning:** The procedure for this type of item was similar to the procedure for *Information* and *Comprehension* items. In this case, there is no special treatment of the beginning of the sentence. After translating to concepts, we removed some very common concepts that proved to be unhelpful The concepts we removed are: 'person', 'get', 'need', 'make', 'out', 'up', 'often', 'look', 'not', 'keep', 'see', and 'come'. For second and third clues, we simply added them to the input into the ConceptNet natural language tools.

**Vocabulary:** We treated these as one-word questions, where the answer had to contain one of the relations *IsA*, *HasA*, *HasProperty*, *UsedFor*, *CapableOf*, *DefinedAs*, *MotivatedByGoal*, or *Causes*.

We introduced the restriction on relations because otherwise the top-scoring answers returned by ConceptNet were not always definitional. For instance, for *house,* the top answer related houses to windows: "Houses usually have windows."

**Similarity:** For similarity items, the inputs were two words, such as "snake" and "alligator". For each word, we identified the concept for the word and its two closest neighbors using the "spreading activation" function of AnalogySpace. For each of those six concepts, we find the 100 highest rated features and their scores. Using AnalogySpace, we created a set of scored predicted features for each word. Each set could have up to 300 entries, though typically both sets had fewer, since we expect many common entries among a concept and its two closest neighbors. We then found the features in the intersection of the two sets, and returned as the answer the highest scored feature. Scores were determined by adding the score from each set.

### 4.2 Determining the VIQ Score

We scored the WPPSI-III subtests once using the top-scored answer to each item, and a second time using the best answer from among the five top-scoring answers to each item. We call the former *strict scoring* and the latter *relaxed.* The comparison between the strict and the relaxed scores gives us information about the robustness of the results.

## 5. Results

### 5.1 VIQ Scores Obtained

Raw scores and scaled scores scaled for a child of age 4 years 0 months, both strict and relaxed, are given for each of the five subtests in Table 1. The raw scores are the number of points earned for items answered correctly, as described at the beginning of Section 3. However, it is the scaled subscores that are important for comparing results between subtests and for comparing to human children's performance. The first observation is that for four of the subtests, the difference between the strict and the relaxed scores is minimal. Only for *Similarities* items are the scores significantly different. The second pertinent observation is that there are huge variations in performance from subtest to subtest. ConceptNet does well on *Vocabulary* and *Similarities*, middling on *Information,* and poorly on *Word Reasoning* and *Comprehension*.

To further compare ConceptNet to human performance, we calculate its overall verbal intelligence score. The calculation consists of two steps. First, a subset of three out of the five verbal subtests are aggregated into a verbal score by adding the individual's subtest scores. There are three variations, illustrated in Table 2: the *standard* score is the sum of Information, Word Reasoning, and Vocabulary, while the *3 best* and *3 worst* versions are self-explanatory. The second step is to enter the aggregated scores into a table based on the norming data that the publishers of WPPSI-III provide along with the test itself. The table yields the VIQ score corresponding to each aggregated raw score.

A special problem that does not arise with the use of the test to measure human performance is to determine the age that should be assumed in reading off the VIQ from the norming table. We calculated the VIQ of ConceptNet under the assumptions that the system is either 4, 5, 6, or 7 years old. That is, we compared ConceptNet's performance to that of four different age cohorts of children.

**Table 1.** Raw WPPSI-III subtest scores and scaled results for a child of age 4 years, 0 months, obtained with the ConceptNet system, using both strict and relaxed scoring regiments (see text). Columns 2-4 show which subtests are used to compute which version of VIQ that is reported in the text and the figures.

| Subtest | Raw Scores | | Scaled Scores (age 4) | |
|---|---|---|---|---|
| | Scoring Regimen | | Scoring Regimen | |
| | Strict | Relaxed | Strict | Relaxed |
| Information | 20 | 21 | 10 | 11 |
| Word Reasoning | 3 | 3 | 7 | 7 |
| Vocabulary | 20 | 21 | 13 | 14 |
| Similarities | 24 | 37 | 13 | 19 |
| Comprehension | 2 | 2 | 7 | 7 |

**Table 2.** Which subtests are used to compute which version of VIQ that is reported in the text and the figures.

|  | Subtests included in VIQ | | |
|---|---|---|---|
| Subtest | Standard | 3 best | 3 worst |
| Information | x | x | x |
| Word Reasoning | x |  | x |
| Vocabulary | x | x |  |
| Similarities |  | x |  |
| Comprehension |  |  | x |

In Figure 2, we show the standard VIQ scores as a function of scoring regimen and the assumed age of the "child." As expected from the inspection of the raw scores, there is little difference between the strict and relaxed scores; that is, the results are robust with respect to scoring method. If we assume that ConceptNet is four years old, its VIQ score is average (VIQ = 100). At an assumed age of five years, the system scores below average with a VIQ of 88. At an assumed age of 7 years, the system scores far below average with a VIQ of 72.

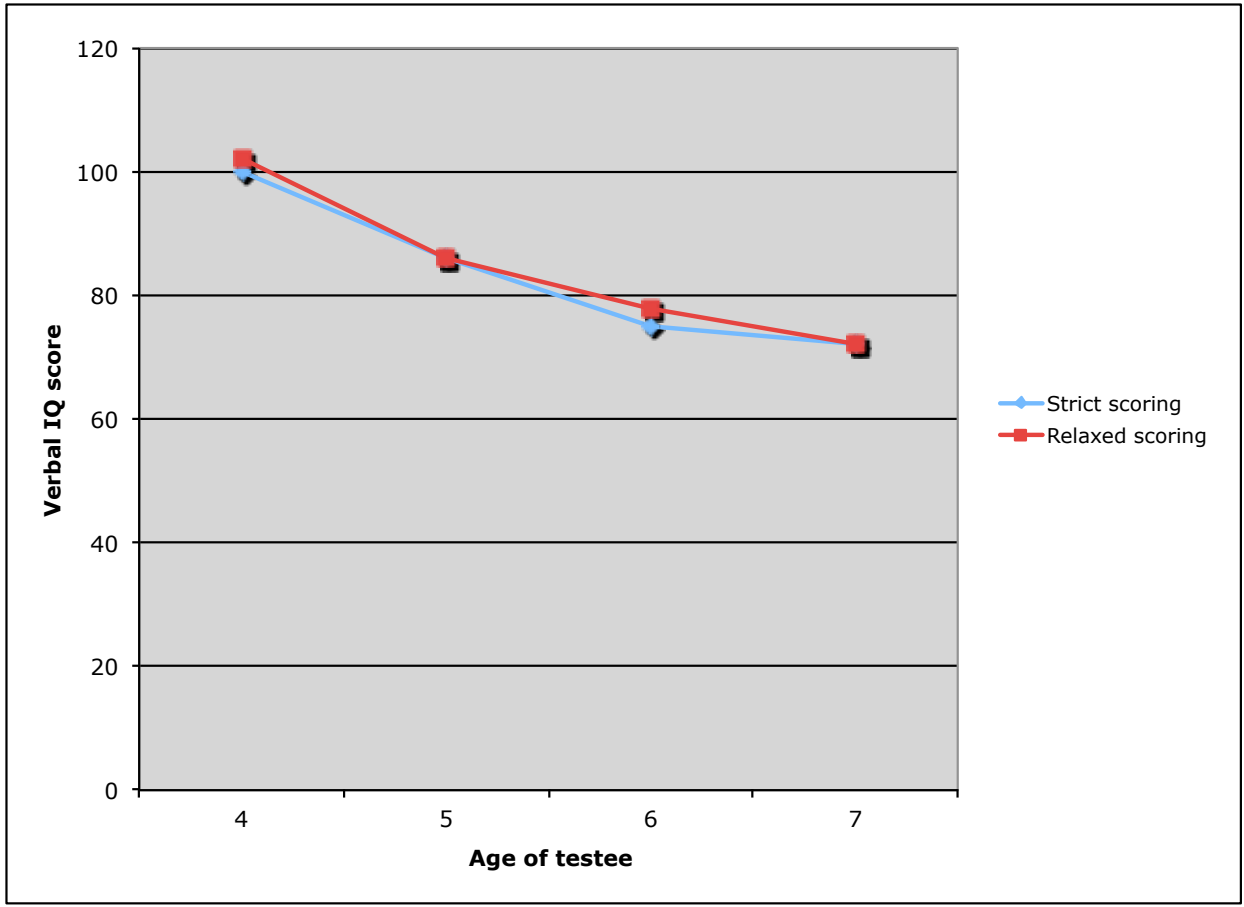

**Fig. 2**. WPPSI-III VIQ of ConceptNet as a function of assumed age in years, computed using the standard test composition, suing both strict and relaxed scoring.

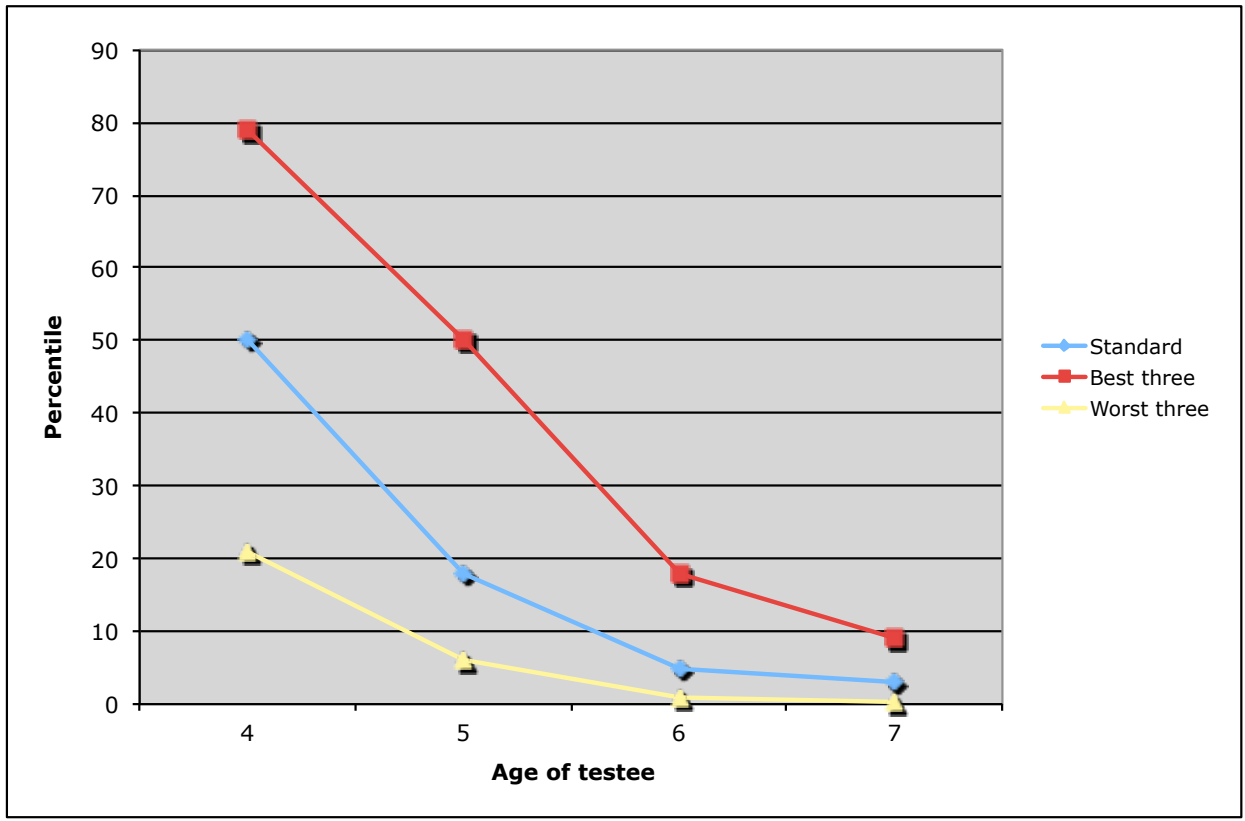

**Fig. 3.** WPPSI-III percentile for VIQ as a function of age, computed using the *3 best, standard*, and *3 worst* test compositions. Strict scoring was used to compute these results.

Transforming the scores into percentiles makes the results clearer. In Figure 3 we show percentiles for the strict scores as a function of age for the standard VIQ variant as well as for the *three best* and *three worst* test compositions. Considered as a 4-year old, the system is in the 21$^{st}$, 50$^{th}$, and 79$^{th}$ percentile (worst, standard, best), while considered as a 7-year old, the system falls below the 10$^{th}$ percentile for all three test variants.

**5.2 Some Qualitative Observations**

We, because of when we happened to begin our work, ran a comparison of two different versions of AnalogySpace/ConceptNet 4. We began our work when the most current release was the February 2010 version, which was updated in March 2012. The March 2012 release was a minor update; the number of concepts in AnalogySpace grew by about 5 percent.

The scores of the two versions of ConceptNet on the WPPSI-III were similar. The only large difference occurred in the Similarity scores. With the earlier version of ConceptNet, the strict Similarity scaled subtest score was 19 (3 standard deviations above the mean) for a 4-year old, and 11 even for a 7-year old. With the later version, the strict scaled subscale for a 4-year old was a still high, but not extraordinary, 13; however, the relaxed score actually went up from the old to new version. So both versions did very well in placing a correct answer somewhere among their top five answers, but evidently the best answer is not necessarily given the highest weight by ConceptNet.

Initially we had hypothesized that Similarity items would be more difficult than either Vocabulary or Information items, because answering Similarity items requires more than mere retrieval. However, as we said, ConceptNet's results on Similarity items were consistently better than its results on Information items, and often better than its results on Vocabulary items. The high score on the Similarities subtest may reflect that abstracting categories is a particular goal of the AnalogySpace designer's use of spectral methods [37].

Results were somewhat sensitive to whether we removed one-word concepts that were part of two-word concepts that AnalogySpace also had. For example, in one method the translation of the Comprehension item "Why do people shake hands?" is to the single concept ['shake hand'] and in the other to the list of three concepts ['shake', 'hand', 'shake hand']. The one-concept query elicits answers of 'thank', 'flirt', and 'meet friend' with relation *HasSubevent*. The three-word version instead gives 'epileptic fit' with relation *HasSubevent* as its top answer. Removing the one-word concepts improved performance considerably on our made-up Comprehension items, and some on the real Comprehension items. In the other direction, oddly, it hurt performance somewhat on our made-up Information items, though it made no significant difference on the WPPSI-III Information items. For example, on our made-up Information item, "Where can you find a teacher?" ['find teacher', 'find', 'teacher'] gives *AtLocation* 'school' as its top answer followed by *AtLocation* 'classroom'. But for ['find teacher'] we get *AtLocation* 'band' followed by *AtLocation* 'piano'. (The scores we report for the WPPSI-III are for the version that *does* remove the one-word concepts for both Information and Comprehension. We committed to that choice before running the WPPSI-III questions because it gave overall better performance on Information and Comprehension questions combined in testing on our made-up items.)

Many wrong answers are not at all like the wrong answers children would give, and seem very much to defy common sense. For example, consider the Word Reasoning item "lion" with the three clues: "This animal has a mane if it is male", "this is an animal that lives in Africa," and "this a big yellowish-brown cat." The five top answers, after all the clues, in order were: dog, farm, creature, home, and cat. Two answers, creature and cat, are in the vague neighborhood of lion. However, the other answers are clear violations of common sense. Common sense should at the very least confine the answer to animals, and should also make the simple inference that, "if the clues say it is a cat, then types of cats are the only alternatives to be considered."

## 6. Discussion

We found that the WPPSI-III VIQ psychometric test gives a WPPSI-III VIQ to ConceptNet 4 that is equivalent to that of an average four-year old. The performance of the system fell when compared to older children, and it compared poorly to seven year olds. This result is far below the 9[th] to 12[th] grade performance claimed by Evans [14] for his visual analogy program. We note that Evans's program was written with the visual analogy test in mind, while we confronted ConceptNet with test items that the creators of the program in all likelihood never considered.

Several factors limited the VIQ obtained. Translating the verbal test items into a form suitable for input to ConceptNet is a complex process that raises multiple issues. For example, ConceptNet does little or no word-sense disambiguation. It combines different forms of one word into one database entry, to increase what is known about that entry. The lack of disambiguation hurts when, for example, the system's natural language processing tools convert *saw* into the base form of the verb *see*, and our question "What is a saw used for?" is answered by "An eye is used to see." The ConceptNet knowledge base does know which is the subject and which is the object in "eye UsedFor see", but the natural language processing tools that take user input to query the system do not make that distinction. In general, more powerful natural language processing tools would likely improve system performance.

Another limitation is the information the system has captured. Some information is missing, as one would expect. Some information is in the very large but less reliable collection of 275,000 concepts, but not in the smaller collection used by the AnalogySpace software. To allow the system the use of the full concept base would likely improve performance.

Another clue to possible areas of improvement is provided by the variations in the subtest scores in Table 1: Performance on *Comprehension* and *Word Reasoning* items was significantly lower than on the other item types. This does not come as a surprise. Answering *why*-questions, which make up most of the *Comprehension* items, is known to be a difficult problem in question answering [24]. Special purpose inference routines for creating explanations and integrating multiple cues would likely raise the system's performance.

One initial attempt at considering the entire large knowledge base and new sorts of routines for query processing from such a knowledge base involves using the tools of network analysis on the large knowledge base [2, 12]. Examples are given for using the analysis of spreading activation to refine the ranking of potential answers, and for using data mining to explore connections between the relations in the network, This, in turn, can be used to identify incorrect or missing entries in the knowledge base.

Future work on the query interface, the knowledge base, and the inference routines will inevitably raise the question of where the boundary is to be drawn between natural language understanding on the one hand, and commonsense reasoning on the other.

Our work highlighted both some picayune issues with ConceptNet 4, such as particular facts not in the knowledge base we used (e.g., the freezing point of water in Fahrenheit), and some genuinely difficult problems, such as answering the "Why" and "What might happen if" questions of the Comprehension subtest as well as a five- or six-year-old child.

More broadly, the areas of commonsense and natural language within AI would benefit by having some benchmarks to drive research over the intermediate term. The Winograd Challenge is one promising attempt to develop such benchmarks [1, 23]. Our use of IQ tests may be another.

In general, recent successes in AI have been mostly *learning driven,* resting upon statistics, large quantities of data, and machine learning. The era of *knowledge driven* AI, resting upon logic, reasoning, and knowledge, appears to have past [10]. Interestingly, each of ConceptNet 4, the recently released ConceptNet 5, Google's knowledge graph, and the output of NELL [7] have at least some aspects of both paradigms. Perhaps knowledge bases that are a hybrid of the two paradigms will play a role in the next round of AI progress.

**Acknowledgments:** Ohlsson was supported, in part, by Award # N00014-09-1-0125 from the Office of Naval Research (ONR), US Navy. Other authors were supported by NSF Grant CCF-0916708.

Documentation on the ConceptNet 4 system, including AnalogySpace and divisi2, is available from http://xnet.media.mit.edu/docs/index.html.

# References

[1] Ackerman, E. 2014. A Better Test than Turing. *IEEE Spectrum*. 51, 10 (2014), 20–21.
[2] Berger-Wolf, T., Diochnos, D.I., London, A., Pluhár, A., Sloan, R.H. and Turán, G. 2013. Commonsense knowledge bases and network analysis. *Logical Formalizations of Commonsense Reasoning* (2013).
[3] Bringsjord, S. 2011. Psychometric artificial intelligence. *Journal of Experimental & Theoretical Artificial Intelligence*. 23, 3 (2011), 271–277.
[4] Bringsjord, S. and Schimanski, B. 2003. What is artificial intelligence? Psychometric AI as an answer. *IJCAI* (2003), 887–893.
[5] Brooks, R. 2008. I, Rodney Brooks, Am a Robot. *IEEE Spectrum*. 45, 6 (2008).
[6] Buchanan, B.G. and Smith, R.G. 1988. Fundamentals of expert systems. *Annual Review of Computer Science*. 3, 1 (1988), 23–58.
[7] Carlson, A., Betteridge, J., Kisiel, B., Settles, B., Hruschka Jr, E.R. and Mitchell, T.M. 2010. Toward an Architecture for Never-Ending Language Learning. *AAAI* (2010).
[8] Cassimatis, N.L. and Bignoli, P. 2011. Microcosms for testing common sense reasoning abilities. *Journal of Experimental & Theoretical Artificial Intelligence*. 23, 3 (2011), 279–298.
[9] Chapin, N., Szymanski, B., Bringsjord, S. and Schimanski, B. 2011. A bottom-up complement to the logic-based top-down approach to the story arrangement test. *Journal of Experimental & Theoretical Artificial Intelligence*. 23, 3 (2011), 329–341.
[10] Cristianini, N. 2014. On the current paradigm in artificial intelligence. *AI Communications*. 27, 1 (Jan. 2014), 37–43.
[11] Davis, E. 1990. *Representations of Commonsense Knowledge*. Morgan Kaufmann Pub.
[12] Diochnos, D.I. 2013. Commonsense Reasoning and Large Network Analysis: A Computational Study of ConceptNet 4. *arXiv:1304.5863 [cs]*. (Apr. 2013).
[13] Dowe, D.L. and Hernández-Orallo, J. 2012. IQ tests are not for machines, yet. *Intelligence*. 40, 2 (2012), 77–81.
[14] Evans, T.G. 1964. A heuristic program to solve geometric-analogy problems. *Proceedings of the April 21-23, 1964, Spring Joint Computer Conference* (1964), 327–338.
[15] Feigenbaum, E.A. 1989. What hath Simon wrought? *Complex information processing: The impact of Herbert A. Simon*. D. Klahr and K. Kotovsky, eds. Psychology Press. 165–182.
[16] Havasi, C., Pustejovsky, J., Speer, R. and Lieberman, H. 2009. Digital intuition: Applying common sense using dimensionality reduction. *Intelligent Systems, IEEE*. 24, 4 (2009), 24–35.
[17] Havasi, C., Speer, R. and Alonso, J. 2007. ConceptNet 3: a flexible, multilingual semantic network for common sense knowledge. *Recent Advances in Natural Language Processing* (2007), 27–29.
[18] Hernández-Orallo, J. and Dowe, D.L. 2010. Measuring universal intelligence: Towards an anytime intelligence test. *Artificial Intelligence*. 174, 18 (2010), 1508–1539.
[19] Hofstadter, D.R. 2008. *Fluid Concepts and Creative Analogies: Computer Models of the Fundamental Mechanisms of Thought*. Basic Books.
[20] Klenk, M., Forbus, K., Tomai, E. and Kim, H. 2011. Using analogical model formulation with sketches to solve Bennett Mechanical Comprehension Test problems. *Journal of Experimental & Theoretical Artificial Intelligence*. 23, 3 (2011), 299–327.
[21] Legg, S. and Hutter, M. 2007. Tests of machine intelligence. *50 years of artificial intelligence*. Springer. 232–242.
[22] Lenat, D.B. 1995. CYC: A large-scale investment in knowledge infrastructure. *Communications of the ACM*. 38, 11 (1995), 33–38.

[23] Levesque, H.J., Davis, E. and Morgenstern, L. 2011. The Winograd Schema Challenge. *AAAI Spring Symposium: Logical Formalizations of Commonsense Reasoning* (2011).

[24] Maybury, M.T. 2004. Question Answering: An Introduction. *New Directions in Question Answering*. 3–18.

[25] McCarthy, J. 1959. Programs with common sense. *Proc. Teddington Conf. on the Mechanization of Thought Processes* (1959).

[26] Mueller, E.T. 2006. *Commonsense Reasoning*. Morgan Kaufmann.

[27] Nilsson, N.J. 2009. *The Quest for Artificial Intelligence*. Cambridge University Press.

[28] Ohlsson, S., Sloan, R.H., Turán, G., Uber, D. and Urasky, A. 2012. An Approach to Evaluate AI Commonsense Reasoning Systems. *Twenty-Fifth International FLAIRS Conference* (2012).

[29] Ohlsson, S., Sloan, R.H., Turán, G. and Urasky, A. 2013. Verbal IQ of a Four-Year Old Achieved by an AI System. *Logical Formalizations of Commonsense Reasoning*. (2013).

[30] Ohlsson, S., Sloan, R.H., Turán, G. and Urasky, A. 2013. Verbal IQ of a Four-Year Old Achieved by an AI System. *Proc. AAAI 2013, Late-Breaking Developments*. (2013).

[31] Perkins, D. 1995. *Outsmarting IQ: The Emerging Science of Learnable Intelligence*. Free Press.

[32] Reitman, W.R. 1965. *Cognition and Thought: An Information Processing Approach*. Wiley.

[33] Russell, S. and Norvig, P. 2009. *Artificial Intelligence: A Modern Approach*. Prentice Hall.

[34] Sanghi, P. and Dowe, D.L. 2003. A computer program capable of passing IQ tests. *Proc. 4th ICCS International Conference on Cognitive Science (ICCS'03)* (2003), 570–575.

[35] Singh, P. 2002. The public acquisition of commonsense knowledge. *Proc.AAAI Spring Symposium on Acquiring (and Using) Linguistic (and World) Knowledge for Information Access* (2002).

[36] Speer, R., Arnold, K. and Havasi, C. 2010. Divisi: Learning from Semantic Networks and Sparse SVD. *9th Python in Science Conference (SCIPY 2010)* (2010).

[37] Speer, R., Havasi, C. and Lieberman, H. 2008. AnalogySpace: Reducing the dimensionality of common sense knowledge. *Proceedings of AAAI* (2008).

[38] Turney, P.D. 2011. Analogy perception applied to seven tests of word comprehension. *Journal of Experimental & Theoretical Artificial Intelligence*. 23, 3 (2011), 343–362.

[39] Veale, T. 2004. WordNet sits the SAT: A Knowledge-Based Approach to Lexical Analogy. *Proc. ECAI'2004, the 16th European Conf. on Artificial Intelligence* (2004).

[40] Wang, H., Gao, B., Bian, J., Tian, F. and Liu, T.-Y. 2015. Solving Verbal Comprehension Questions in IQ Test by Knowledge-Powered Word Embedding. *arXiv:1505.07909 [cs.CL]*. (Jul. 2015).

[41] Weston, J., Bordes, A., Chopra, S., Mikolov, T. and Rush, A.M. 2015. Towards AI-Complete Question Answering: A Set of Prerequisite Toy Tasks. *arXiv:1502.05698 [cs, stat]*. (Feb. 2015).